\documentclass[10pt,twocolumn,letterpaper]{article}
\usepackage{cvpr}             
\usepackage[dvipsnames]{xcolor}

\definecolor{cvprblue}{rgb}{0.21,0.49,0.74}
\usepackage[pagebackref,breaklinks,colorlinks,citecolor=cvprblue]{hyperref}
\usepackage[accsupp]{axessibility} 
\usepackage{threeparttable}
\usepackage{multirow}
\usepackage{tabularx}
\usepackage{ragged2e}
\usepackage{booktabs}
\usepackage{makecell}
\usepackage{xspace}
\usepackage{diagbox}
\usepackage{xcolor}
\usepackage{soul}
\usepackage{amssymb}
\usepackage{comment}
	
\usepackage{color, colortbl}

\usepackage{tikz}
\usetikzlibrary{fit,calc}

\colorlet{pink}{red!40}

\usepackage{algorithm}
\usepackage{algpseudocode}
\usepackage{ragged2e}
\usepackage{hyperref}
\usepackage{xspace}

\newcolumntype{C}[1]{>{\centering\arraybackslash}p{#1}}
\newcolumntype{P}[1]{>{\centering\arraybackslash}p{#1}}
\newcolumntype{M}[1]{>{\centering\arraybackslash}m{#1}}
\newcolumntype{L}[1]{>{\raggedright\arraybackslash}p{#1}}
\newcolumntype{R}[1]{>{\raggedleft\arraybackslash}p{#1}}
\newcolumntype{J}[1]{>{\justifying\arraybackslash}p{#1}}

\usepackage{tikz}
\usepackage{comment}

\newcommand{\papersource}[1]{\scriptsize{\color{blue}{#1}}}
\newcommand{\mypara}[1]{\noindent\textbf{#1}}

\newcommand{\myMethod}{\mbox{CoHFF}\xspace}

\definecolor{green_size}{RGB}{23, 156, 125}
\definecolor{purple_number}{RGB}{112, 48, 160}
\makeatletter
\def\blfootnote{\xdef\@thefnmark{}\@footnotetext}
\makeatother

\title{Collaborative Semantic Occupancy Prediction \\with Hybrid Feature Fusion in Connected Automated Vehicles}

\author{
    Rui Song\textsuperscript{\rm 1,\rm 2 *},
    Chenwei Liang\textsuperscript{\rm 1},
    Hu Cao\textsuperscript{\rm 2},
    Zhiran Yan\textsuperscript{\rm 3},
    Walter Zimmer\textsuperscript{\rm 2},\\
    \vspace{1ex}
    Markus Gross\textsuperscript{\rm 1},
    Andreas Festag\textsuperscript{\rm 1,\rm 3},
    Alois Knoll\textsuperscript{\rm 2}\\
    \vspace{1ex}
    $^{1}$Fraunhofer IVI \quad $^{2}$Technical University of Munich \quad $^{3}$Technische Hochschule Ingolstadt\\
    \url{https://rruisong.github.io/publications/CoHFF}\\
}

\begin{document}
\twocolumn[{
\renewcommand\twocolumn[1][]{#1}%
\maketitle
\begin{center}
    \centering
    \captionsetup{type=figure}
    \includegraphics[trim={0.5cm 0cm 0.2cm 0cm},clip, width=1\textwidth]{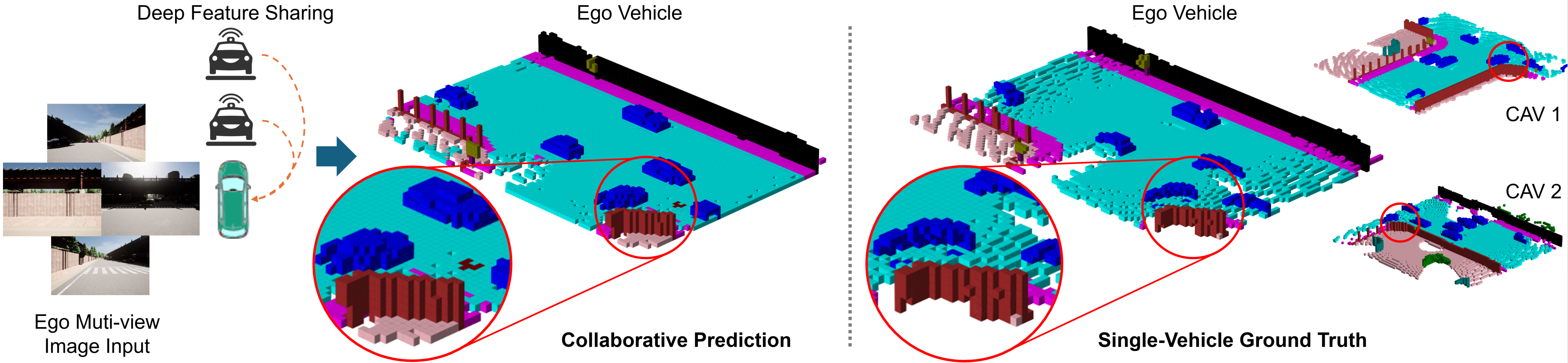}
    \captionof{figure}{Collaborative semantic occupancy prediction leverages the power of collaboration in multi-agent systems for 3D occupancy prediction and semantic segmentation. This approach enables a deeper understanding of the 3D road environment by sharing latent features among connected automated vehicles (CAVs), surpassing the ground truth captured by a multi-camera system in the ego vehicle.}
    \label{fig:teaser}
\end{center}
}]

\maketitle
\blfootnote{*Corresponding author, email address: rui.song@ivi.fraunhofer.de}

\begin{abstract}
Collaborative perception in automated vehicles leverages the exchange of information between agents, aiming to elevate perception results.
Previous camera-based collaborative 3D perception methods typically employ 3D bounding boxes or bird's eye views as representations of the environment.
However, these approaches fall short in offering a comprehensive 3D environmental prediction.
To bridge this gap, we introduce the first method for collaborative 3D semantic occupancy prediction.
Particularly, it improves local 3D semantic occupancy predictions by hybrid fusion of (i) semantic and occupancy task features, and (ii) compressed orthogonal attention features shared between vehicles.
Additionally, due to the lack of a collaborative perception dataset designed for semantic occupancy prediction, we augment a current collaborative perception dataset to include 3D collaborative semantic occupancy labels for a more robust evaluation.
The experimental findings highlight that: (i) our collaborative semantic occupancy predictions excel above the results from single vehicles by over 30\%, and (ii) models anchored on semantic occupancy outpace state-of-the-art collaborative 3D detection techniques in subsequent perception applications, showcasing enhanced accuracy and enriched semantic-awareness in road environments. 

\end{abstract}

\section{Introduction}

Collaborative perception, also known as cooperative perception, significantly improves the accuracy and completeness of each connected and automated vehicle's (CAV) sensing capabilities by integrating multiple viewpoints, surpassing the limitations of single-vehicle systems~\cite{hu2022where2comm, liu2020who2com, liu2020when2com, hu2023collaboration, xu2022v2x, yu2023vehicle, li2021learning, wang2020v2vnet, Yang_2023_ICCV, li2023multi}. This approach enables CAVs to achieve comparable or superior perception abilities, even with cost-effective sensors. Notably, recent research in~\cite{hu2023collaboration} suggests that camera-based systems may outperform LiDAR in 3D perception through collaboration in Vehicle-to-Everything (V2X) communication networks. Previous studies in camera-based collaborative perception typically processed inputs from various CAVs into simplified formats such as 3D bounding boxes or Bird's Eye View (BEV) segmentation. While efficient, these methods tend to miss important 3D semantic details, which are indispensable for holistic scene understanding and reliable execution of downstream applications.

Lately, camera-based 3D semantic occupancy prediction, also known as semantic scene completion~\cite{roldao20223d}, has become a pioneering method in 3D perception~ {\cite{cao2022monoscene,fang2023tbp,ganesh2023octran,huang2023tri,jiang2023symphonize,li2023voxformer,liu2023towards,miao2023occdepth,min2023occ,tan2023ovo,tong2023scene,wang2023panoocc,wang2023pet,wei2023surroundocc,zhang2023bev,zhang2023occformer, Yao_2023_ICCV}}. This approach uses RGB camera data to predict the semantic occupancy status of voxels in 3D space, involving both the determination of voxel occupancy and semantic classes of occupied voxels. This research enhances single CAVs' environmental understanding, improving decision-making in downstream applications for automated vehicles. However, this task based on RGB imagery through collaborative methods has not been explored.

 To bridge this gap, we delve into the feasibility of 3D semantic occupancy prediction in the context of collaborative perception, as shown in Fig.~\ref{fig:teaser}, and introduce the Collaborative Hybrid Feature Fusion (CoHFF) Framework. Our approach involves separate pre-training for the dual subtasks of predicting both semantics and occupancy. We then extract the high-dimensional features from these pretrained models for dual fusion processes: inter-CAV semantic information fusion via V2X Feature Fusion, and intra-CAV fusion of semantic information with occupancy status through task feature fusion. This fusion yields a comprehensive decoding of each voxel's occupancy and semantic details in 3D space. \newline

In order to evaluate the performance of our framework, we extend the existing collaborative perception dataset OPV2V~\cite{xu2022opv2v}. By reproducing OPV2V scenarios in the CARLA simulator, we collect comprehensive 3D voxel groundtruth with semantic labels across 12 categories. Our experiments show, that for the task of semantic occupancy prediction, a collaborative approach significantly outperforms single-vehicle performance in most categories, as intuitively expected. We also validate the effectiveness of task feature fusion: our findings show that the task fusion, by incorporating features as prior knowledge of each other, enhances subtask performance beyond what separately trained models achieved. Additionally, training tasks independently result in more task-specific features and thus can be easier to compress. Our experiments prove that we achieve more complex 3D perception with a communication volume comparable to existing methods.

\begin{figure*}[t!]
   \centering
   \includegraphics[width=1\textwidth]{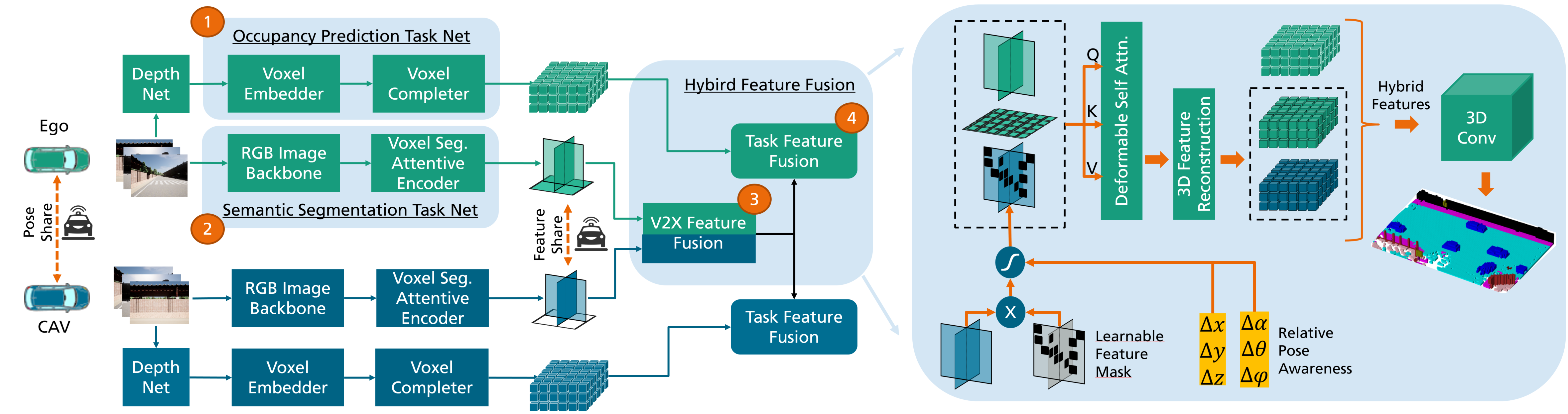}
   \caption{The CoHFF Framework consists of four key modules: (1) Occupancy Prediction Task Net, for occupancy feature extraction; (2) Semantic Segmentation Task Net, creating semantic plane-based embeddings; (3) V2X Feature Fusion, merging CAV features via deformable self-attention; and (4) Task Feature Fusion, uniting all task features to enhance semantic occupancy prediction.}
   \label{fig:system}
\end{figure*}

\noindent\textbf{Contributions} 
To summarize, our main contributions are threefold:
\begin{itemize}
\item We introduce the first camera-based framework for collaborative semantic occupancy prediction, enabling more precise and comprehensive 3D semantic occupancy segmentation than single-vehicle systems through feature sharing  in V2X communication networks. The performance can be enhanced by over 30\% via collaboration.
\item We propose the hybrid feature fusion approach, which not only facilitates efficient collaboration among CAVs, but also markedly enhances the performance over models pre-trained solely for occupancy prediction or semantic voxel segmentation.
\item We enrich the collaborative perception dataset OPV2V~\cite{xu2022opv2v} with voxel ground truth containing 12~categories semantic, bolstering the framework evaluation. Our method, \myMethod, achieves comparable results to current leading methods in subsequent 3D perception applications, and additionally offers more semantic details in road environment. 
\end{itemize}

\section{Related work}

\subsection{Collaborative perception}

In intelligent transportation systems, collaborative perception empowers CAVs to attain a more accurate and holistic understanding of the road environment via V2X communication and data fusion. Typically, data fusion in collaborative perception falls into three categories: early, middle, and late fusion. Given the bandwidth limitations of V2X networks, the prevalent approach is middle fusion, where deep latent space features are exchanged~\cite{hu2022where2comm, liu2020who2com, liu2020when2com, hu2023collaboration, xu2022v2x, yu2023vehicle, li2021learning, wang2020v2vnet, Yang_2023_ICCV, li2023multi}. The advantage of middle fusion lies in its ability to convey critical information beyond mere object-level details, bypassing the need to share raw data. The development of datasets specifically designed for collaborative perception~\cite{li2022v2x, hu2022where2comm, yu2022dair, yu2023v2x, xu2023v2v4real, hao2024rcooper, zimmer2024tumtraf, ma2024holovic} 
has led to remarkable progress in learning-based approaches in recent years. However, these datasets fall short in offering ground truth data for 3D semantic occupancy, which motivates us to extend the dataset in this work, aiming to access the performance of collaborative semantic occupancy prediction.

\mypara{Collaborative Camera 3D Perception}. Compared to LiDAR-driven collaborative perception~\cite{xu2023v2v4real}, camera-based methods are often more challenging, due to the absence of explicit depth information in RGB data. However, given the lower price and smaller weight of cameras, they inherently have a higher potential for large-scale deployment. Previous work in~\cite{xu2022cobevt} and~\cite{hu2023collaboration} has validated that, with collaboration, camera-based 3D perception can match or even outperform LiDAR performance. Given that current research on camera-based collaborative perception either focuses on 3D bounding box detection and BEV semantic segmentation, there remains a research gap in semantic occupancy prediction. Hence, in this study, our aim is to pioneer and explore the topic of collaborative occupancy segmentation.

\subsection{Camera-based semantic occupancy prediction}

Occupancy segmentation, which segments a voxel-based 3D environment model~\cite{mescheder2019occupancy,zhou2018voxelnet}, has achieved notable success in the realm of autonomous driving. Original occupancy segmentation methods lean heavily on LiDAR, since its point cloud inherits 3D information, aligning naturally with voxel-based environmental models. The recent work proposed in~\cite{li2023multi} explored the collaborative semantic occupancy prediction based on LiDAR. However, with cameras offering richer environmental details, camera-driven 3D occupancy segmentation is gradually emerging as a novel domain. Recent work in the past year, e.g.~{\cite{cao2022monoscene,fang2023tbp,ganesh2023octran,huang2023tri,jiang2023symphonize,li2023voxformer,liu2023towards,miao2023occdepth,min2023occ,tan2023ovo,tong2023scene,wang2023panoocc,wang2023pet,wei2023surroundocc,zhang2023bev,zhang2023occformer,li2023fb,hayler2023s4c}} have also delved into methods for achieving semantic occupancy prediction based on RGB data, yielding promising performance, but only for single vehicle perception.

Furthermore, the datasets for the vision-based 3D Semantic Occupancy Prediction, e.g. Semantic-KITTI~\cite{behley2021ijrr}, SSC-Benchmark~\cite{li2023sscbench}, OpenOccupancy~\cite{wang2023openoccupancy}, and Occ3D~\cite{tian2023occ3d} have been developed specifically for camera-based 3D occupancy segmentation tasks, thus offering resources for continued research. However, those datasets do not support collaborative perception in multi-agent scenarios. Generally, agents sharing different perspective information through collaboration can further enhance voxel-based occupancy segmentation. Due to semantic occupancy prediction offering a more nuanced 3D environmental understanding than collaborative 3D perception methods focused on bounding boxes or BEV perception, it likely requires the exchange of more complex, higher-dimensional features. Determining the most effective information for communication to facilitate the transmission of denser, more informative data stands as a significant challenge.

\subsection{Plane-based features}
TPVFormer~\cite{huang2023tri} decomposes features for occupancy segmentation into a 3D space. \cite{fridovich2023k} introduced a K-Planes decomposition technique designed to reconstruct static 3D scenes and dynamic 4D videos. Building on the foundations laid by~\cite{fridovich2023k}, and drawing inspiration from~\cite{huang2023tri}, we consider to project semantically relevant information onto orthogonal planes, facilitating information sharing through more streamlined communication. By sharing these plane-based features, we establish the foundational structure of our approach.

\section{Methodology}
Our CoHFF framework consists of four key modules, namely occupancy prediction Task Net, Semantic Segmentation Task Net, V2X Feature Fusion and Task Feature Fusion, as shown in Fig.~\ref{fig:system}. It achieves camera-based collaborative semantic occupancy prediction by sharing plane-based semantic features via V2X communication.
\subsection{Problem formulation}
\label{sec:3.1}
Given a network of CAVs, defined by a global communication network represented as an undirect graph $\mathcal{G} = (\mathcal{N}, \mathcal{E})$. For each CAV $i$, the set of connected CAVs, is denoted by $\mathcal{N}_{i} = \{j|(i, j) \in \mathcal{E}\}$, where $\mathcal{E}$ is the existing communication links between two CAVs, and $j$ denotes the index of the CAVs connecting to $i$. We consider the input data in RGB format, and denote $\mathcal{I}_i$ as the image data for a CAV $i$. The environment model is represented as a 3D voxel grid in one hot embedding $\mathbf{V} \in \mathbb{R}^{X\times Y\times Z\times C}$, where $X$, $Y$ and $Z$ are voxel grid dimensions. For each CAV $i$, $\mathbf{V}_i$ represents the predicted occupancy of voxels, while $\mathbf{V}^{(0)}_i$ represents the ground truth of these voxels. The objective of collaborative semantic occupancy prediction, as aligned with the optimization problem in~\cite{hu2022where2comm, hu2023collaboration}, is defined as follows:
\begin{multline}
    \max_{\theta,M} \sum_i g(\Phi_\theta (\mathcal{I}_i, \{\mathcal{M}_{i\rightarrow j}|j\in\mathcal{N}_{i}\}), \mathbf{V}_i^{(0)}), \\
    s.t. \sum_i |\{\mathcal{M}_{i\rightarrow j}|j\in\mathcal{N}_{i}\}|\leq B,
    \label{eq:cp}
\end{multline}
where $g(\cdot)$ is the perception metric for optimization.
$\Phi$ represents the model parametrized by $\theta$, and $\mathcal{M}_{i\rightarrow j}$ denotes the message transmitted from CAV $i$ to CAV $j$. The size of these messages is constrained by a communication budget upper bound $B\in \mathbb{R}^+$. 

Considering the communication upper bound, instead of directly sending high-dimensional voxel-sized features $\mathcal{F}^{V}$, we opt to transmit features $\mathcal{F}^{\mathbf{P}}$ from orthogonal planes. This approach reduces the messages from $\mathcal{M}^{\mathcal{F}^{V}} \in \mathbb{R}^{X \times Y \times Z \times F}$ to $\mathcal{M}^{\mathbf{P}^{xz}} \in \mathbb{R}^{X \times Z \times F}$ and $\mathcal{M}^{\mathbf{P}^{yz}} \in \mathbb{R}^{Y \times Z\times F}$, where $\mathbf{P}^{xz}$ and $\mathbf{P}^{yz}$ denote the features projected on the $xz$- and $yz$-planes respectively. $F$ represents the length of a single feature vector. For instance, in a voxel space of ${100\,\times\,100\,\times\,8}$ with a feature dimension of $128$, transmitting orthogonal plane features can reduce communication volume by $50\,\times$, from 39.05\,MB to 0.78\,MB, which is comparable to existing collaborative perception methods, yet it offers more extensive and detailed semantic 3D scene information. Based on these considerations, we introduce our framework in the following section.

\subsection{Framework design}

We divide our method into two distinct pre-communication tasks: 3D occupancy prediction and semantic voxel segmentation.
We believe that occupancy features enhance the semantic segmentation performance by providing geometry insight of distinct object classes. Meanwhile, semantic information can suggest changes of a voxel occupancy. Based on this interplay, our approach initially focuses on independent pre-training for each task. Then we fuse the features from both tasks to learn a combined semantic occupancy predictor that yields better performance for each individual task. This assumption is experimentally validated by the ablation study in~Tab.~\ref{table:comm}.
Consequently, our framework comprises two specialized pre-trained networks: an occupancy prediction task network and a semantic segmentation task network, as shown in Fig.~\ref{fig:system}.

\mypara{Occupancy prediction task network}. 
The occupancy prediction necessitates the conversion of 2D image data into a 3D occupancy grid.
We first use an off-the-shelf depth prediction network $\Phi^{depth}(\cdot)$ to determine the depth of each pixel. Following the work in~\cite{hu2022where2comm, hu2023collaboration}, we employ CaDNN~\cite{CaDDN} for depth estimation. This depth data is then embedded into voxel space through a 3D Emedder, resulting in a preliminary voxel representation. This voxel-based road environment is further completed by a 3D occupancy encoder $\Phi^{occ}(\cdot)$. Finally, the occupancy task features $\mathbf{F}^{occ} \in \mathbb{R}^{X \times Y \times Z \times F}$ is extracted for task fusion.

\mypara{Semantic segmentation task network}.
In the segmentation network, we process RGB data to generate feature maps $\mathbf{F}^{seg}$ using $\Phi^{img}(\cdot)$, which are then subjected to deformable cross-attention~\cite{zhu2021deformable} to facilitate mapping onto a 3D semantic segmentation space. 
Drawing inspiration from K-Planes~\cite{fridovich2023k} and TPVformer~\cite{huang2023tri}, we project these features onto three spatially orthogonal planes $\mathcal{P}=\{\mathbf{P}^{xy},\mathbf{P}^{xz},\mathbf{P}^{yz}\}$.
Among these dense and informative 3D feature representations, two are transmitted via V2X messages, i.e. $\mathcal{M} = \{\mathcal{M}^{\mathbf{P}^{xz}}, \mathcal{M}^{\mathbf{P}^{yz}} \}$. The reason behind not sending the $\mathbf{P}^{xy}$ plane, is that the we use the $\mathbf{P}^{xy}$ of the ego vehicle for reconstructing the 3D features, which facilitates the alignment of the feature space with the detection range of interest of ego vehicle.

Both networks generate high-dimensional features that are fed into a hybrid feature fusion network, thereby forming the core of CoHFF for semantic occupancy prediction. 

\subsection{Hybrid feature fusion}

\mypara{V2X Feature Fusion}. Given one CAV $j$ communicating to the ego vehicle $i$, the features of the CAV condensed by the segmentation network can contain overlapping information, particularly regarding semantics in proximity to the ego vehicle, which the ego vehicle itself can accurately predict. We implement a masking technique to selectively filter these plane-based features of the CAV, before they are communicated to the ego vehicle. By adjusting a sparsification rate hyperparameter, we reduce the volume of the CAV´s plane-based features shared during collaboration, in line with the communication budget. The compressed message $\bar{\mathcal{M}} = \{\bar{\mathcal{M}}^{\mathbf{P}^{xz}}, \bar{\mathcal{M}}^{\mathbf{P}^{yz}} \}$ can be acquired as follows: 
\begin{equation}
\bar{\mathbf{P}}_j^{xz},\bar{\mathbf{P}}_j^{yz} \leftarrow \mathbf{P}_j^{xz} \odot \mathbf{H}_j^{xz} ,\mathbf{P}_j^{yz} \odot\mathbf{H}_j^{yz},
\end{equation}
where $\mathbf{H}_j^{xz}$ and $\mathbf{H}_j^{yz}$ represent the learnable feature masks for features on x-z and y-z planes.

Additionally, we ensure relative pose awareness between the ego vehicle and other CAVs. Specifically, we feed the filtered plane features and the relative pose information into an MLP network combined with a Sigmoid function, in line with the methodology proposed in~\cite{liu2023petrv2}.

We now attend these pose-aware filtered plane features from the CAV ($\bar{\mathbf{P}}_j^{xz}$, $\bar{\mathbf{P}}_j^{yz}$) over the three plane features of the ego vehicle ($\mathbf{P}_i^{xy}$, $\mathbf{P}_i^{xz}$, $\mathbf{P}_i^{yz}$). 
In particular, we use deformable self-attention to update the all five feature planes.  
The fusion and updating of these planes are accomplished by plane self-attention ($PSA$), as follows:
\begin{equation}
    PSA(\mathbf{p}) = DA(\mathbf{p},\mathcal{R},\{\mathbf{P}_i, \bar{\mathbf{P}}_j^{xz}, \bar{\mathbf{P}}_j^{yz}|j\in \mathcal{N}_i\} ),
    \label{eq:psa_fusion}
\end{equation}
where $DA(\cdot)$ is deformable self-attention, $\mathbf{p}\in\mathbb{R}^F$ is a query and $\mathcal{R}$ is a set of reference points, as described in~\cite{zhu2021deformable}. $\mathbf{P}_i$ denotes all the three planes in ego vehicle. 

The updated 2D plane features are used in the next step to reconstruct 3D semantic segmentation features $\mathbf{F}^{seg}$. The semantic segmentation feature $\mathbf{f}_{x,y,z}^{seg}$ at a specific Voxel location $x,y,z$ can be reconstructed as follows:
\begin{equation}
    \mathbf{f}_{x,y,z}^{seg} = \mathbf{p}^{xy}_{i,z} + \bar{\mathbf{p}}^{xz}_{j,y} + \bar{\mathbf{p}}^{yz}_{j,x} \in \mathbb{R}^{F}, \forall j \in \mathcal{N}_i,
    \label{eq:expand}
\end{equation}
where $ \bar{\mathbf{p}}^{xz}_{j,y}$ and $\bar{\mathbf{p}}^{yz}_{j,x}$ is plane features from CAV $j$, and $\mathbf{p}^{xy}_{i,z}$ is the plane (BEV) features from ego vehicle. This idea of sum of projected features for 3D reconstruction is originally proposed in~\cite{huang2023tri}, with our work adapting it to multi-agent scenarios. 

\mypara{Task Feature Fusion}. After retrieving global semantic information as $\mathbf{F}^{seg}$, the final step aims at fusion with features $\mathbf{F}^{occ}$ from the occupancy prediction task. To accomplish this, $\mathbf{F}^{seg}$ and $\mathbf{F}^{occ}$ are concatenated and passed to a 3D depth-wise convolution network~\cite{ye20193d}, in order to produce the final semantic voxel map. This task feature fusion network $\Phi^{tff}(\cdot)$ is implemented as follows:
\begin{equation}
    \mathbf{V_i} = \Phi^{tff}(\mathbf{F}_i^{occ}, \mathbf{F}_i^{seg}, \{\mathbf{F}_j^{seg}|j\in\mathcal{N}_i\})\in \mathbb{R}^{X \times Y \times Z \times C}.
\end{equation}
The CoHFF pseudocode is given in Algorithm~\ref{alg:cohff}. 

\begin{algorithm}[t!]
\centering
 \caption{\raggedright: CoHFF framework for collaborative semantic occupancy prediction.}
 \label{alg:cohff}
 \begin{algorithmic}[1]
 \newcommand{\algorithmicbreak}{\textbf{break}}
 \newcommand{\BREAK}{\State \algorithmicbreak} 
\renewcommand{\algorithmicrequire}{\textbf{Input:}}
 \renewcommand{\algorithmicensure}{\textbf{Output:}}
    \For{each CAV $i$ \textbf{in parallel}}
    \State $\mathbf{F}_i^{occ} \leftarrow \Phi^{occ}(Proj(\Phi^{depth}(\mathcal{I}_i), \mathcal{I}_i)))$
    \State $\mathbf{F}_i^{img}\leftarrow \Phi^{img}(\mathcal{I}_i) $
    \State update plane-based features $\mathbf{P}_i^{xz},\mathbf{P}_i^{yz}, \mathbf{P}_i^{xy}$ using deformable cross- and self-attention~\cite{zhu2021deformable} 
    \State $\bar{\mathbf{P}}_i^{xz},\bar{\mathbf{P}}_i^{yz} \leftarrow \mathbf{P}_i^{xz} \odot \mathbf{H}_i^{xz} ,\mathbf{P}_i^{yz} \odot\mathbf{H}_i^{yz} $
    \State $\bar{\mathcal{M}_i} \leftarrow \{\bar{\mathbf{P}}_i^{xz} , \bar{\mathbf{P}}_i^{yz} \}$
    \State CAV $i$ broadcasts messages $\bar{\mathcal{M}_i}$
        \For{$j \in \mathcal{N}_i$}
            \State CAV $i$ receives messages $\bar{\mathcal{M}_j}$
        \EndFor
    \State update $\{\mathbf{P}_i, \bar{\mathbf{P}}_j^{xz}, \bar{\mathbf{P}}_j^{yz}|j\in \mathcal{N}_i\}$ using self-attention based on (\ref{eq:psa_fusion}) 
    \State reconstruct $F_j^{seg}$ based on (\ref{eq:expand}) \Comment{VFF}
    \State $\mathbf{V_i} \leftarrow \Phi^{tff}(\mathbf{F}_i^{occ}, \mathbf{F}_i^{seg}, \{\mathbf{F}_j^{seg}|j\in\mathcal{N}_i\})$ \Comment{TFF}
    \EndFor  
\end{algorithmic}
\end{algorithm}

\subsection{Losses}

We train the completion network training using focal loss proposed in~\cite{Lin_2017_ICCV}, applying it to a dataset with binary labels $\{0, 1\}$. 
For both the segmentation network and the hybrid feature fusion network, we employ a weighted cross-entropy loss to train for semantic labels. 
Notably, in this context, the label for the \emph{empty} is also designated as 0.

\section{Dataset}

To effectively evaluate collaborative semantic occupancy prediction, a dataset that supports collaborative perception and includes 3D semantic occupancy labels is crucial. 
Thus, we enhance the OPV2V dataset~\cite{xu2022opv2v} by integrating 12 different 3D semantic occupancy labels, as shown in Tab.~\ref{table:comparison}
This enhancement is achieved using the high-fidelity CARLA simulator~\cite{dosovitskiy2017carla} and the OpenCDA autonomous driving simulation framework~\cite{xu2023opencda}. We position four semantic LiDARs at the original camera sites to precisely capture the semantic occupancy ground truth within the cameras' FoV. In addition, we associate ground truth data from all CAVs to create a detailed collaborative ground truth for collaborative supervision. Furthermore, to comprehensively capture occluded semantic occupancies for all CAVs, we include a simulation replay in our data collection process, where each CAV is equipped with 18 semantic LiDARs. This strategic configuration is crucial for effectively evaluating completion tasks, as it guarantees extensive data collection, encompassing areas not visible in direct associated FoV. In alignment with the original OPV2V protocol, we replay the simulation and generate a multi-tier ground truth.

\section{Experimental evaluation}

\setlength{\tabcolsep}{3pt}
\begin{table}[t!]
\centering
\fontsize{9}{12}\selectfont
\begin{threeparttable}
\caption{Comparison 3D object detection with AP\tnote{2} of vehicles.}
\label{table:comparison_3d}
\begin{tabular}
{L{3.2cm}|C{1.8cm}|C{1.1cm}C{1.1cm}}
\toprule
Approach  &\# Agents &  AP@0.5 & AP@0.7  \\
\midrule
\midrule
DiscoNet~\papersource{(NeurIPS~21)}   & Up to 7 &   36.00 & 12.50  \\
V2X-ViT~\papersource{(ECCV~22)}  & Up to 7 & 39.82 & 16.43   \\
Where2Comm~\papersource{(NeurIPS~22)}  & Up to 7& 47.30 & 19.30 \\
CoCa3D~\papersource{(CVPR~23)}  & 7\tnote{1}& \textbf{69.10} & \textbf{49.50} \\
\myMethod  & Up to 7 & 48.51 & 36.39 \\
\midrule
CoCa3D-2~\papersource{(CVPR~23)} &2& 25.90 & 12.60 \\
\myMethod &2& ~\textbf{36.63}  & \textbf{27.95}\\
\bottomrule 
\end{tabular}
\begin{tablenotes}
    \scriptsize 
    \item[1] CoCa3D is trained on OPV2V+, where extended agents provide more input information for better results.
    \item[2] We calculate the 3D IoU by comparing the predicted voxels with the ground truth voxels for each object, rather than using 3D bounding boxes due to the potential unnecessary occupancy in 3D bounding boxes.
   \end{tablenotes}
\end{threeparttable}
\end{table}

\setlength{\tabcolsep}{3pt}
\begin{table}[t!]
\centering
\fontsize{9}{12}\selectfont
\resizebox{\columnwidth}{!}{%
\begin{threeparttable}
\caption{Comparison of BEV semantic segmentation with IoU in the class of Vehicle, Road and Others. }
\label{table:comparison_bev}
\begin{tabular}
{L{2.5cm}|C{1.8cm}|C{1.1cm}C{1.1cm}C{1.1cm}}
\toprule
Approach  & \# Agents   & \rotatebox[origin=c]{0}{Vehicle}  & \rotatebox[origin=c]{0}{Road} & \rotatebox[origin=c]{0}{Others\tnote{1}}\\
\midrule
\midrule
CoBEVT~\papersource{(CoRL~22)} & 2    & 46.13 & 52.41 & - \\
\myMethod  &  2   & \textbf{47.40} &  \textbf{63.36} & \textbf{40.27}\\
\midrule
CoBEVT~\papersource{(CoRL~22)} & Up to 7   &  60.40 & \textbf{63.00} & -\\
\myMethod &  Up to 7  & \textbf{64.44} & 57.28 & \textbf{45.89} \\ 
\bottomrule 
\end{tabular}
\begin{tablenotes}
    \scriptsize 
    \item[1] It refers to additional object classes identified through semantic segmentation predictions projected onto the BEV plane. These categories include buildings, fences, terrain, poles, vegetation, walls, guard rails, traffic signs, and bridges. The IoU for these objects is calculated and reported as IoU.
\end{tablenotes}
\end{threeparttable}
}
\end{table}

\setlength{\tabcolsep}{3pt}
\begin{table}[t!]
\centering
\fontsize{9}{12}\selectfont
\begin{threeparttable}
\caption{CoHFF achieves robust IoU and mIoU performance, when the communication volume (CV) is reduced by setting various sparsification rates (Spar. Rate). }
\label{table:comm}
\begin{tabular}
{L{2cm}|C{1cm}C{1cm}C{1cm}C{1cm}C{1cm}}
\toprule
Spar. Rate & 0.00  & 0.50 & 0.80 & 0.95 & 0.99\\
\midrule
\midrule
CV (MB) ($\downarrow$) & 16.53 & 8.27 & 3.31 & 0.83 & 0.17\\
\midrule
IoU ($\uparrow$) & 50.46& 49.56 & 49.53 & 48.52 & 48.02\\
mIoU ($\uparrow$) & 34.16  & 32.97 & 32.70 & 30.13 & 29.48\\
\bottomrule 
\end{tabular}

\end{threeparttable}
\end{table}

\setlength{\tabcolsep}{3pt}
\begin{table*}[t!]
\centering
\fontsize{9}{12}\selectfont
\begin{threeparttable}
\caption{Component ablation study on occupancy prediction (Occ. Pred.), semantic segmentation (Sem. Seg.), and semantic occupancy prediction (Sem. Occ. Pred.) tasks. The components include:  Occupancy Prediction Task Net (OPTN), Semantic Segmentation Task Net (SSTN), Task Feature Fusion (TFF) and V2X Feature Fusion (VFF). The gray color in table cells indicates that the corresponding component is not applicable for the task.}
\label{table:comparison}
\begin{tabular}
{R{3.3cm}||C{1cm}C{1cm}C{1cm}C{1cm}||C{1cm}C{1cm}C{1cm}||C{1cm}C{1cm}}
\toprule
\centering{Task type} &  \multicolumn{4}{c||}{Occ. Pred.} & \multicolumn{3}{c||}{Sem. Seg.} & \multicolumn{2}{c}{Sem. Occ. Pred.} \\
\midrule
OPTN&RL~\tnote{1} & \checkmark& \checkmark&\checkmark&\cellcolor{gray!25}&\cellcolor{gray!25}&\cellcolor{gray!25}&\checkmark&\checkmark\\
SSTN &\cellcolor{gray!25}&\cellcolor{gray!25}&\cellcolor{gray!25}&\cellcolor{gray!25}&\checkmark&\checkmark&\checkmark&\checkmark&\checkmark\\
TFF&&&\checkmark&\checkmark&&\checkmark&\checkmark&\checkmark&\checkmark\\
VFF (Collaboration)&&&&\checkmark&&&\checkmark&&\checkmark\\
\midrule
 IoU ($\uparrow$)&49.35&67.22&76.62 &\textbf{86.37}&41.30&42.11&\textbf{51.38} &38.52&\textbf{50.46}\\
 mIoU ($\uparrow$)&57.12&64.01& 59.16&\textbf{69.15}&21.59&30.51&\textbf{35.91} &24.85&\textbf{34.16}\\
\midrule
Building (5.40\%)~~\fcolorbox{black}{red}{\rule{0pt}{6pt}\rule{6pt}{0pt}}&67.50&\textbf{68.36}&41.29&48.41&9.65&\textbf{27.25}&15.06&21.04&\textbf{25.72}\\
Fence (0.85\%)~~\fcolorbox{black}{brown}{\rule{0pt}{6pt}\rule{6pt}{0pt}}&59.40&62.05& 51.60&\textbf{65.01}&11.67&30.29&\textbf{30.91}&20.50&\textbf{27.83}\\
Terrain (4.80\%)~~\fcolorbox{black}{pink}{\rule{0pt}{6pt}\rule{6pt}{0pt}}&43.60&49.78&68.21 &\textbf{79.81}&51.18&51.41&\textbf{61.98}&43.93&\textbf{48.30}\\
Pole (0.39\%)~~\fcolorbox{black}{yellow}{\rule{0pt}{6pt}\rule{6pt}{0pt}}&66.30&\textbf{70.67}& 62.31&64.12&2.14&36.80&\textbf{40.74}&31.66&\textbf{42.74}\\
Road (40.53\%)~~\fcolorbox{black}{cyan}{\rule{0pt}{6pt}\rule{6pt}{0pt}}&51.47&77.78& 91.26&\textbf{93.00}&56.82&60.02&\textbf{64.09}& 55.83&\textbf{61.77}\\
Side walk (35.64\%)~~\fcolorbox{black}{magenta}{\rule{0pt}{6pt}\rule{6pt}{0pt}}&45.46&58.46& 74.37&\textbf{90.53}&25.22&16.87&\textbf{36.03}&17.31&\textbf{39.62}\\
Vegetation (1.11\%)~~\fcolorbox{black}{green}{\rule{0pt}{6pt}\rule{6pt}{0pt}}&43.61&\textbf{44.43}& 38.87&41.57&9.12&\textbf{22.13}&20.99&14.49&\textbf{20.59}\\
Vehicles (9.14\%)~~\fcolorbox{black}{blue}{\rule{0pt}{6pt}\rule{6pt}{0pt}}&41.40&63.53& 59.52&\textbf{76.48}&59.58&69.81&\textbf{75.88}&58.55&\textbf{63.28}\\
Wall (2.01\%)~~\fcolorbox{black}{black}{\rule{0pt}{6pt}\rule{6pt}{0pt}}&71.51&79.35&49.63 &\textbf{81.20}&32.55&39.80&\textbf{58.49}&33.30&\textbf{58.27}\\
Guard rail (0.04\%)~~\fcolorbox{black}[HTML]{8C368C}{\rule{0pt}{6pt}\rule{6pt}{0pt}}&\textbf{49.67}&46.03&41.35 &43.33&1.10&\textbf{1.95}&1.80&1.54&\textbf{1.94}\\
Traffic signs (0.05\%)~~\fcolorbox{black}{olive}{\rule{0pt}{6pt}\rule{6pt}{0pt}}&68.98&\textbf{69.41}&52.35 &62.54&0.00&9.77&\textbf{11.69}&0.00&\textbf{16.33}\\
Bridge (0.04\%)~~\fcolorbox{black}{teal}{\rule{0pt}{6pt}\rule{6pt}{0pt}}&76.53&78.23& 79.08&\textbf{83.84}&0.00&0.00&\textbf{13.30}&0.00&\textbf{3.53}\\
\bottomrule
\end{tabular}
\begin{tablenotes}
    \scriptsize 
    \item[1]  RL (Raw LiDAR) is used as a baseline for the evaluation on the task of occupancy prediction.
\end{tablenotes}
\end{threeparttable}
\end{table*} 

\subsection{Experiment setup}

\mypara{Baselines}.
Considering the unexplored domain of collaborative occupancy segmentation, we extend the findings from \myMethod to address downstream applications, including BEV perception and 3D detection. In our analysis, we evaluate these outcomes with those from state-of-the-art collaborative perception models that employ multi-view cameras: CoBEVT~\cite{xu2022cobevt} for BEV perception and CoCa3D~\cite{hu2023collaboration} for 3D detection. Furthermore, we examine contemporary methods that integrate alternative modalities, particularly those blending LiDAR with camera inputs or relying solely on LiDAR, including DiscoNet~\cite{li2021learning}, V2X-ViT~\cite{xu2022v2x} and Where2Comm~\cite{hu2022where2comm}. 

\mypara{Implementation details}.
Following the previous work for collaborative perception evaluation on the OPV2V dataset used in~\cite{hu2022where2comm}, we utilize a $40\times40\times 3.2$ meter detection area with a grid size of 100\,\,x\,\,100\,\,x\,\,8, resulting in a voxel size of $0.4~m^3$. We allow CAVs to transmit and share features with a length of 128 for V2X Feature Fusion. Our experiment incorporates the analysis of 12~semantic labels plus an additional \emph{empty} label. We employ CaDNN~\cite{CaDDN} with 50 depth categories and a single out-of-range category for depth estimation, as well as ResNet101~\cite{he2016deep} and FPN~\cite{lin2017feature} as RGB the image backbone. For Voxel completion, we utilize a 3D depth-wise CNN~\cite{ye20193d} and use deformable attention~\cite{zhu2021deformable} in hybrid feature fusion.

\mypara{Evaluation metrics}.
Following the evaluation of semantic occupancy prediction in previous work, such as~\cite{cao2022monoscene, huang2023tri, li2023voxformer}, we primarily utilize the metric Intersection over Union (IoU) for evaluation. This involves calculating IoU for each individual class and the mean IoU (mIoU) across all classes. Additionally, for evaluations in subsequent applications, we compute the Average Precision (AP) at IoU threshold of 0.5 and 0.7, and BEV 2D IoU to compare with other baselines. Specifically, the AP value is calculated only for voxels labeled as vehicles, and the IoU is determined for each pair of predicted and actual vehicles. For BEV IoU, voxels are projected onto the BEV plane and categorized into the corresponding semantic classes.

\subsection{Comparison}

\mypara{Collaborative 3D object detection}. 
First, we compare the performance of CoHFF in 3D detection applications. As shown in Tab.~\ref{table:comparison_3d}, with up to 7 agents' collaborative perception, CoHFF achieves comparable performance to Where2comm at AP@0.5 and obtains an 88.5\% improvement at AP@0.7. We believe this is primarily due to semantic occupancy prediction, which makes the perception results closer to the actual observed shapes, rather than inferring a non-existent bounding box in the scenarios. We also observe that CoCa3D, on the OPV2V+ dataset~\cite{hu2023collaboration},
achieves significantly better performance due to receiving more information from CAVs. To compare directly with CoCa3D, we also conduct scenarios where only two agents communicated at a time. We can see that CoHFF has made significant improvements at both AP@0.5 and AP@0.7.

\mypara{Collaborative BEV segementation}.
Tab.~\ref{table:comparison_bev} presents a comparison between CoHFF and CoBEVT in BEV  semantic segmentation. Note that errors in height prediction from 3D voxel occupancy mapping to the BEV plane may be overlooked during the projection process. Despite this, CoHFF achieves even better performance in predicting vehicles and roads in BEV compared to CoBEVT. Additionally, CoHFF is capable of detecting a wider range of other semantic categories in 3D occupancy.

\subsection{Ablation study}

To validate our hypothesis that independently obtained semantic and occupancy feature information can simultaneously strengthen the original semantic and occupancy tasks, we have decomposed the semantic occupancy prediction into two separate tasks. Tab.~\ref{table:comparison} shows an ablation study by altering the components used. Meanwhile, we also verify the enhancement of collaborative perception over single vehicle perception in terms of semantic occupancy.

\mypara{CoHFF for occupancy prediction}.
When focusing solely on binary occupancy predictions (as shown at Occ. Pred. in Tab.~\ref{table:comparison}), we use voxels processed from raw LiDAR point clouds as a reference, and analyze the IoU in different semantic classes based on semantic occupancy in ground truth. It is observed that by utilizing an occupancy prediction task network to process depth predictions, the overall prediction accuracy is enhanced. Additionally, significant improvements in predicting large objects in occupancy results are noted by integrating features from a semantic segmentation task network, leading to an increased overall IoU. However, a concurrent decline in the mIoU is observed alongside the increase in IoU. This phenomenon is attributed to the influence of semantic features, which seem to steer the model towards prioritizing easily detectable categories, potentially at the expense of smaller or less distinct categories. Finally, through collaboration, the overall IoU and mIoU are further strengthened on the basis of task feature fusion.

\mypara{CoHFF for semantic segmentation}.
In our semantic segmentation task (as shown at Sem. Seg. in Tab.~\ref{table:comparison}), after integrating features from occupancy prediction, we observe an approximate 2\% increase in IoU, but a more substantial over 41\% enhancement in mIoU. We attribute this improvement to the features derived from occupancy prediction, which seem to aid the easier detection of smaller-scale objects, thereby refining their semantic predictions. Consistent with the occupancy prediction task, the final collaboration further elevates the results of semantic segmentation.

\mypara{Collaboration enhances semantic occupancy prediction}.
In the final evaluation of our semantic occupancy prediction (see column Sem. Occ. Pred. in Tab.~\ref{table:comparison}), we further demonstrate the benefits brought by collaboration. By collaboration, the IoU for each category is improved. Notably, some previously undetectable, low-prevalence categories such as traffic signs and bridges can be detected after collaboration. Ultimately, there is an approximate \textbf{31\%} increase in overall IoU and around a \textbf{37\%} enhancement in mIoU.

\subsection{Robustness with low communication budget}
In Tab.~\ref{table:comm}, we increase the sparsification rate to mask the plane-based features transmitted by CAVs, achieving efficient V2X information exchange under a low communication budget. The CoHFF model exhibits stable IoU performance across various levels of sparsification. Even when the communication volume is shrinked by $97\,\times$, the accuracy only decreases by 5\% compared to the original. Meanwhile, the mIoU drops by 15\%. Despite this, due to the model's training under collaborative supervision, it still outperforms the non-collaborative approach.

\subsection{Visual analysis}

\begin{figure*}[t!]
   \centering
   \includegraphics[trim={0.5cm 0 0 0},clip, width=0.98\textwidth]{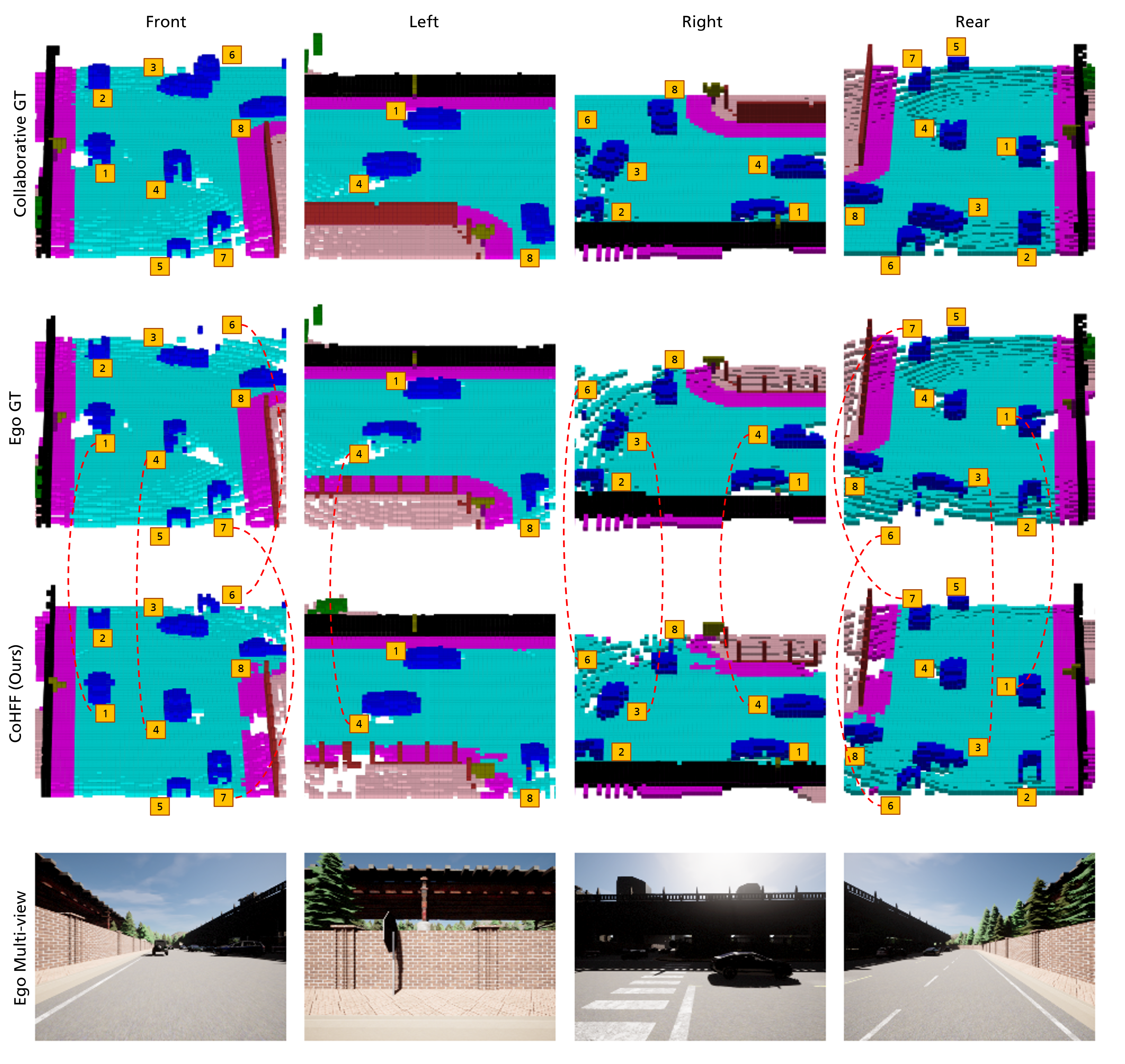}
   \caption{Illustration of collaborative semantic occupancy prediction from multiple perspectives, compared to the ground truth in the ego vehicle's FoV and the collaborative FoV across CAVs. This visualization emphasizes the advanced object detection capabilities in collaborative settings, particularly for objects obscured in the ego vehicle's FoV, such as the vehicle with ID~6.}
   \label{fig:visual_analysis}
\end{figure*}

Fig.~\ref{fig:visual_analysis} presents visual results from the CoHFF model, which are compared from multiple perspectives with the ground truth data, i.e. the ground truth in the ego vehicle's FoV (Ego GT) and the ground truth across all CAVs FoVs (Collaborative GT). It is evident that, overall, the model accurately predicts voxels in various classes such as roads, sidewalks, traffic signs, walls, and fences. We particularly focus on vehicle predictions, as they are among the most critical categories in road environment perception. For clarity, each vehicle object in the figure is numbered.

\mypara{Vehicle geometry completion}. The CoHFF model predicts more complete vehicle objects than those in the Ego GT, such as vehicles 1, 3, 4, and 7. In some instances, the predictions even surpass the completeness of vehicle shapes found in Collaborative GT.

\mypara{Occluded vehicle detection}. CoHFF successfully predicts vehicles outside of the FoV, such as vehicle~6, by utilizing minimal pixel information. This demonstrates that CoHFF can effectively detect occluded vehicles.

\section{Conclusion}
In this work, we explore the task of camera-based semantic occupancy prediction through the lens of collaborative perception. We introduce the CoHFF framework, which significantly enhances the perception performance by over 30\% through integrating features from different tasks and various CAVs. Since currently no dataset specifically designed for collaborative semantic occupancy prediction exists, we also extend the OPV2V dataset with 3D semantic occupancy labels. Our experiments validate that collaboration yields better semantic occupancy prediction results than single-vehicle approaches.

\mypara{Limitation}. Although we demonstrate the immense potential of collaboration for semantic occupancy prediction using simulation data, its performance with real-world data remains to be verified. The collection and development of a specialized dataset, repleted with semantic occupancy labels and derived from multi-agent perception scenarios in real-world settings, are highly  anticipated.

\section{Acknowledgements}
This work was supported by the German Federal Ministry for Digital
and Transport (BMVI) in the project ”5GoIng – 5G Innovation Concept Ingolstadt”.

{
    \small
    \bibliographystyle{ieeenat_fullname}
    \bibliography{main}
}

\onecolumn
\newpage
\twocolumn
\maketitlesupplementary
In this supplementary material, we provide more details of Semantic-OPV2V in Sec.~\ref{sec:supp_dataset}. To showcase the robustness of the proposed CoHFF approach, we give extended results for its performance in relation to the communication budget and additionally assess its robustness in the presence of GPS noise in Sec.~\ref{sec:supp_robustness}. We also present a range of visual results illustrating the effectiveness of CoHFF in diverse scenarios in Sec.~\ref{sec:supp_visual}.  
Note that we consistently use the same color scheme for each semantic class, as illustrated in the first column in Tab.~\ref{table:comm_2}.
\appendix
\section{Semantic-OPV2V dataset}
\label{sec:supp_dataset}

\begin{figure}[ht!]
   \centering
   \includegraphics[width=0.5\textwidth]{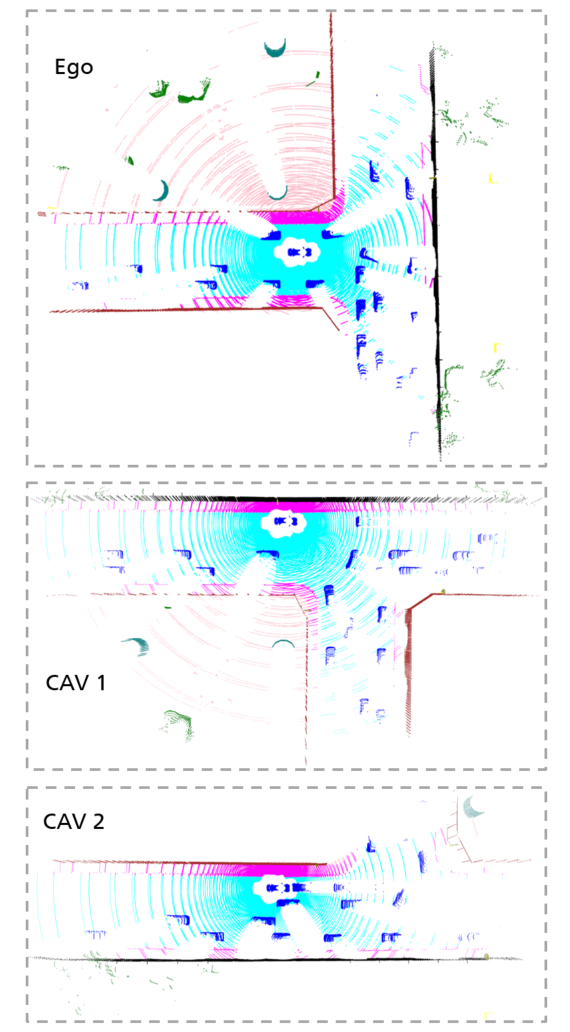}
   \caption{Visualization of semantic point clouds from 4 semantic LiDARs in Ego vehicle and CAVs.}
   \label{fig:4lidar}
\end{figure}

\begin{figure}[ht!]
   \centering
   \includegraphics[width=0.5\textwidth]{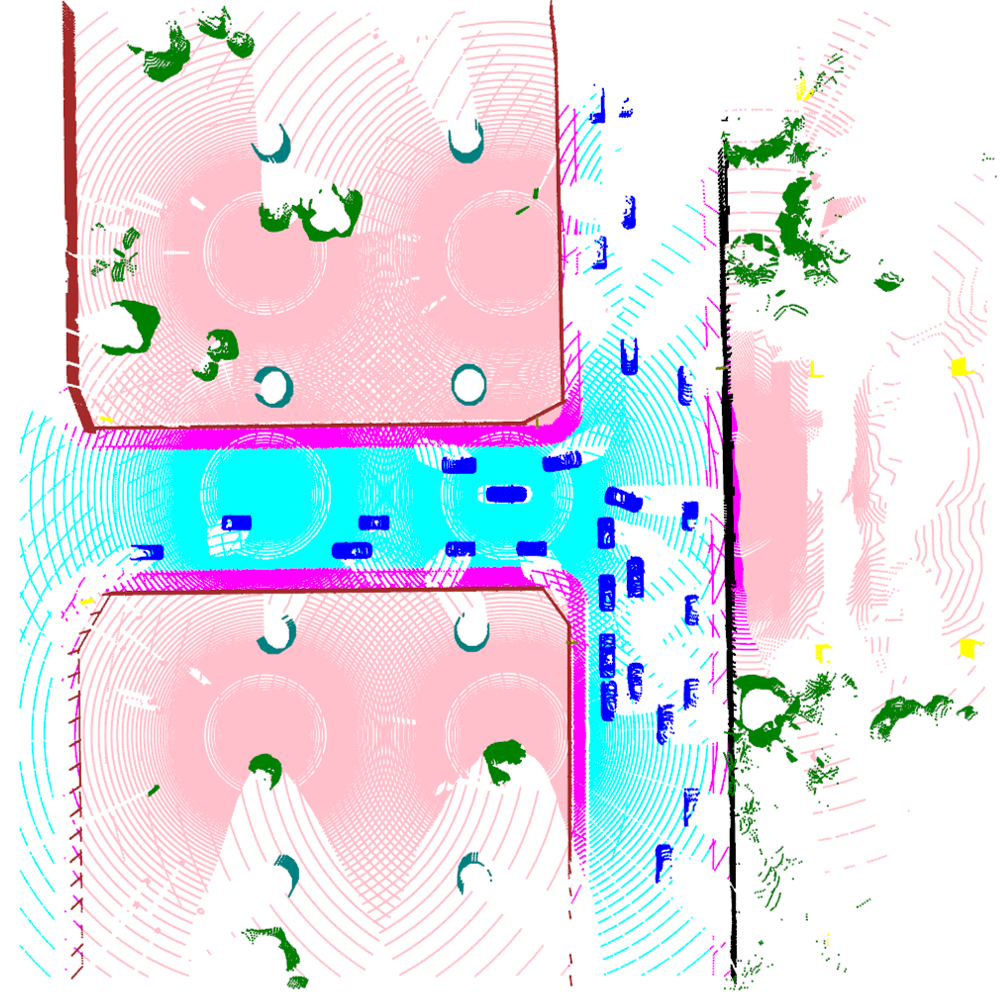}
   \caption{Visualization of semantic point clouds from 18 semantic LiDARs.}
   \label{fig:18lidar}
\end{figure}

We first equip each Connected and Automated Vehicle (CAV) in the CARLA simulation~\cite{dosovitskiy2017carla} with a semantic LiDAR at the position of each camera. This setup aims to capture the road environment within the Field of View (FoV) of the cameras as comprehensively as possible. Fig.~\ref{fig:4lidar} illustrates the semantically labeled point clouds generated by these semantic LiDARs.

Additionally, we outfit the surroundings of each CAV with a system comprising 18 semantic LiDARs to collect data on the road environment, including semantic occupancy space with occluded objects, as shown in Fig.~\ref{fig:18lidar}. Specifically, we choose 9 positions surrounding each CAV, with each adjacent position spaced 30 meters apart. At each of these positions, we install two semantic LiDARs: one set at an vertical FoV ranging from -20 to -90 degrees, and the other ranging from -20 to 0 degrees.

By replaying the OPV2V dataset in CARLA-based OpenCDA~\cite{xu2023opencda}, we collect semantically-labeled point clouds with 4 and 18 semantic LiDARs for each frame in the dataset. These point clouds are saved in PCD-format for further processing into semantic voxel data, useful for supervision or evaluation purposes.

\begin{figure*}[t!]
   \centering
   \includegraphics[width=1\textwidth]{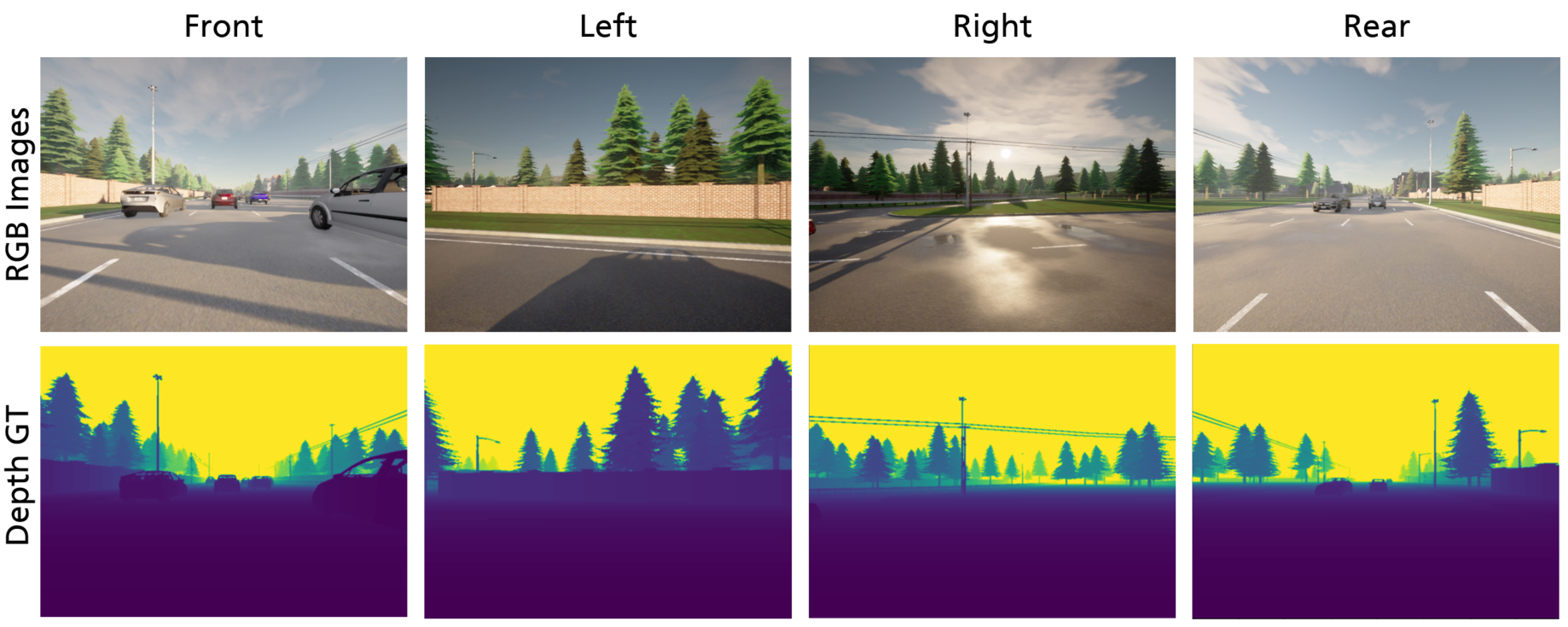}
   \caption{Corresponding depth labels gathered for the RGB cameras in the training dataset.}
   \label{fig:depth}
\end{figure*}

\begin{figure}[t!]
   \centering
   \includegraphics[width=0.5\textwidth]{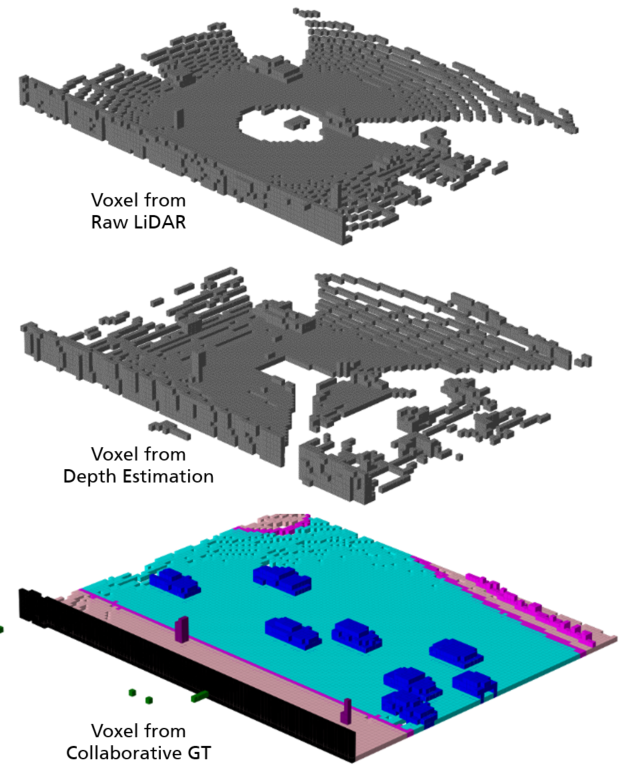}
   \caption{Visual comparison of occupied voxels derived from depth estimation, raw LiDAR, and collaborative semantic voxel labels. The gray color represents occupied voxels with an unknown semantic label.}
   \label{fig:voxel_comparison}
\end{figure}

Moreover, to train the Depth Net, we gather corresponding depth labels for the RGB cameras in the training dataset, as shown in Fig.~\ref{fig:depth}. For a visual evaluation, we transform and visualize the results of depth estimation in the 3D voxel space. Fig.~\ref{fig:voxel_comparison} compares these results with voxels based on raw LiDAR and collaborative semantic voxel labels.

\section{Robustness}
\label{sec:supp_robustness}

\setlength{\tabcolsep}{3pt}
\begin{table}[t!]
\centering
\fontsize{9}{12}\selectfont
\begin{threeparttable}
\caption{CoHFF achieves robust IoU and mIoU performance, when the communication volume (CV) is reduced by setting various sparsification rates (Spar. Rate). The mask used for sparsification is learned under collaborative supervision.}
\label{table:comm_2}
\begin{tabular}
{R{2.3cm}|C{1cm}C{1cm}C{1cm}C{1cm}C{1cm}}
\toprule
Spar. Rate & 0.00  & 0.50 & 0.80 & 0.95 & 0.99\\
\midrule
\midrule
CV (MB) ($\downarrow$) & 16.53 & 8.27 & 3.31 & 0.83 & 0.17\\
\midrule
IoU ($\uparrow$) & 50.46& 49.56 & 49.53 & 48.52 & 48.02\\
mIoU ($\uparrow$) & 34.16  & 32.97 & 32.70 & 30.13 & 29.48\\
\midrule
Building~~\fcolorbox{black}{red}{\rule{0pt}{6pt}\rule{6pt}{0pt}}&25.72&17.77&16.79&13.08&12.12\\
Fence~~\fcolorbox{black}{brown}{\rule{0pt}{6pt}\rule{6pt}{0pt}}&27.83&29.61&29.12&25.25&22.76\\
Terrain~~\fcolorbox{black}{pink}{\rule{0pt}{6pt}\rule{6pt}{0pt}}&48.30&47.98&47.60&44.42&44.77\\
Pole~~\fcolorbox{black}{yellow}{\rule{0pt}{6pt}\rule{6pt}{0pt}}&42.74&37.73&37.69&35.65&35.83\\
Road~~\fcolorbox{black}{cyan}{\rule{0pt}{6pt}\rule{6pt}{0pt}}&61.77&59.47&60.15&59.42&59.86\\
Side walk~~\fcolorbox{black}{magenta}{\rule{0pt}{6pt}\rule{6pt}{0pt}}&39.62&42.03&41.36&40.81&39.11\\
Vegetation~~\fcolorbox{black}{green}{\rule{0pt}{6pt}\rule{6pt}{0pt}}&20.59&21.36&20.18&13.35&14.74\\
Vehicles~~\fcolorbox{black}{blue}{\rule{0pt}{6pt}\rule{6pt}{0pt}}&63.28&60.25&60.33&60.14&59.98\\
Wall~~\fcolorbox{black}{black}{\rule{0pt}{6pt}\rule{6pt}{0pt}}&58.27&52.68&53.41&51.94&51.20\\
Guard rail~~\fcolorbox{black}[HTML]{8C368C}{\rule{0pt}{6pt}\rule{6pt}{0pt}}&1.94&3.86&3.51&1.66&1.55\\
Traffic signs~~\fcolorbox{black}{olive}{\rule{0pt}{6pt}\rule{6pt}{0pt}}&16.33&19.50&19.09&13.13&10.74\\
Bridge~~\fcolorbox{black}{teal}{\rule{0pt}{6pt}\rule{6pt}{0pt}}&3.53&3.39&3.11&2.67&1.11\\
\bottomrule
\end{tabular}

\end{threeparttable}
\end{table}

\subsection{Low communication budget}

We present additional results of the CoHFF performance in reducing the communication budget in Tab.~\ref{table:comm_2}. This includes an assessment of the robust performance for overall Intersection over Union (IoU) as well as individual IoU for each class.

\subsection{GPS noise}

\begin{figure}[t!]
   \centering
   \includegraphics[width=0.5\textwidth]{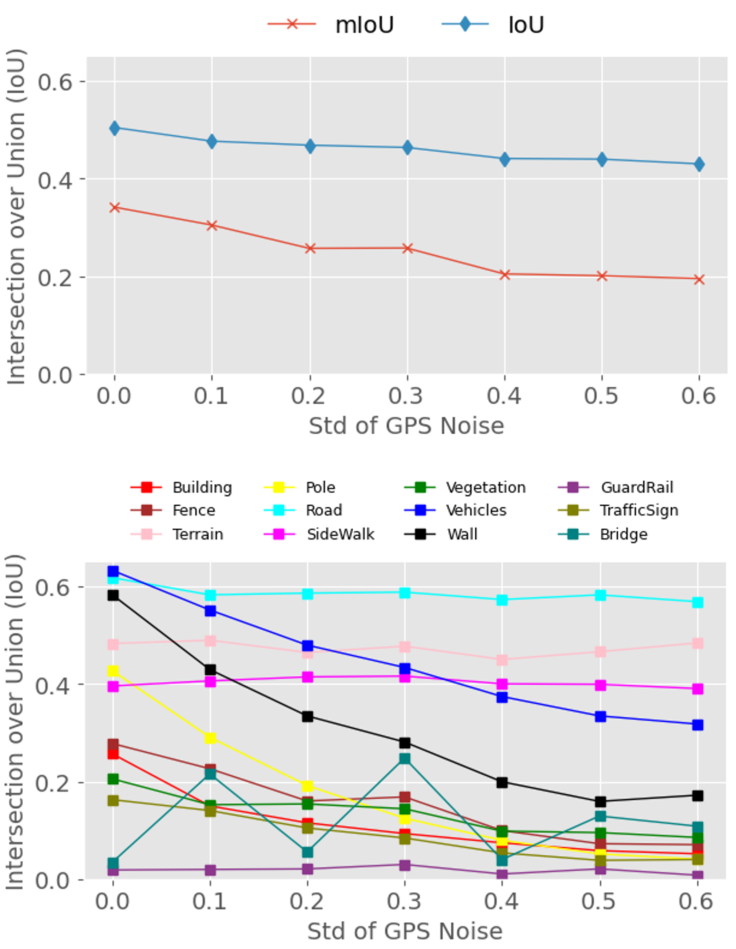}
   \caption{Our CoHFF model demonstrates robust performance in terms of overall IoU and mIoU stability. However, the IoU for each class exhibits individual variations, reflecting the unique impact of GPS noise on different categories. }
   \label{fig:gpsnoise}
\end{figure}

In our paper, we assess the performance of CoHFF using accurate GPS information. This section extends the experiment to include scenarios with varying GPS noise levels in Fig.~\ref{fig:gpsnoise}, specifically Gaussian noise with a standard deviation ranging from 0~m to 0.6~m, which aligns with methodologies used in previous work, such as~\cite{hu2022where2comm, xu2022v2x, hu2023collaboration} for evaluating collaborative perception.

\section{Further visual results}
\label{sec:supp_visual}
\begin{figure*}[t!]
   \centering
   \includegraphics[trim={0 0 0 0},clip, width=0.98\textwidth]{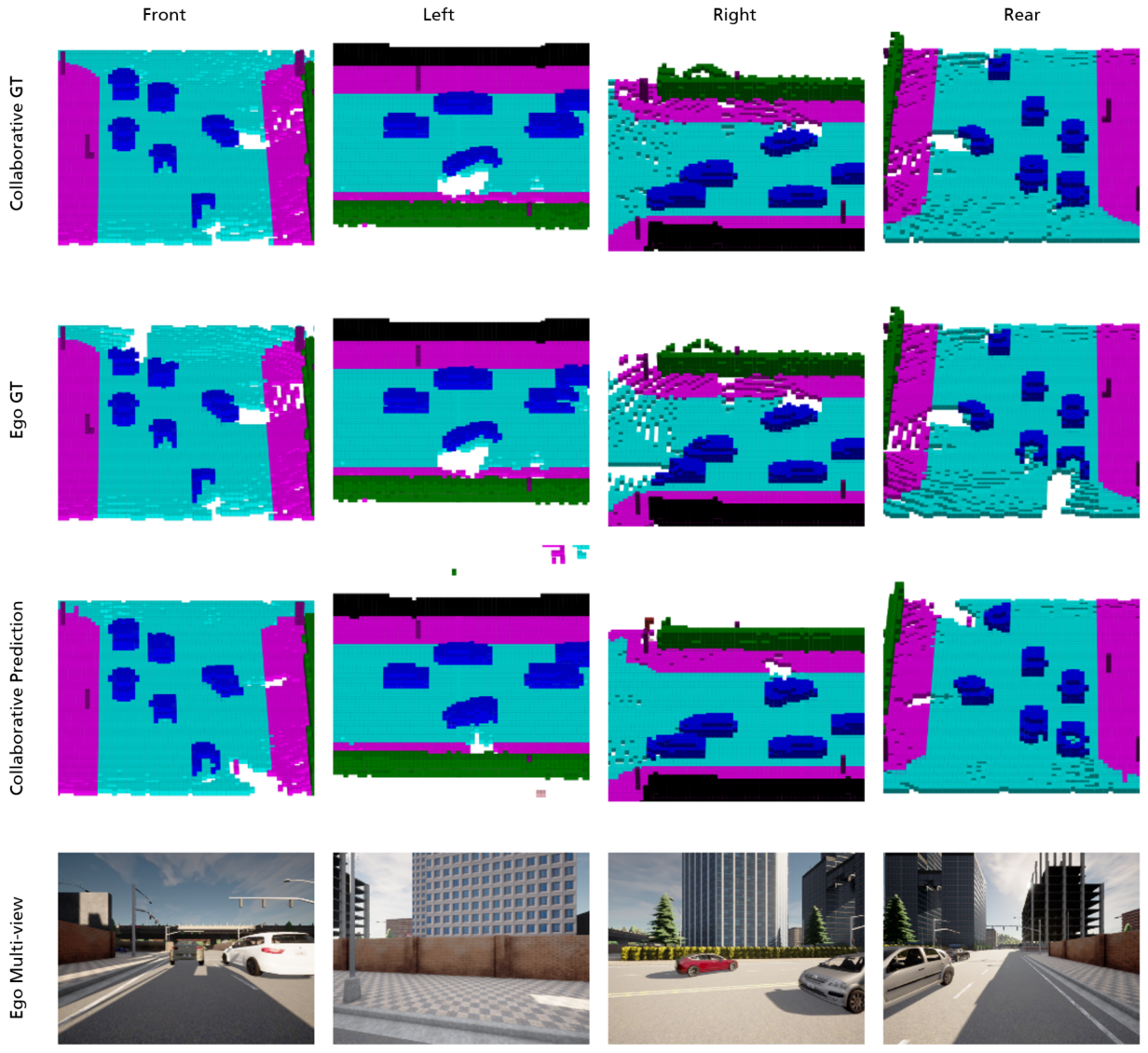}
   \caption{Visual comparison of CoHFF prediction results with collaborative and ego GT in an urban lane-change scenario.}
   \label{fig:visual_analysis_1}
\end{figure*}

\begin{figure*}[t!]
   \centering
   \includegraphics[trim={0 0 0 0},clip, width=0.98\textwidth]{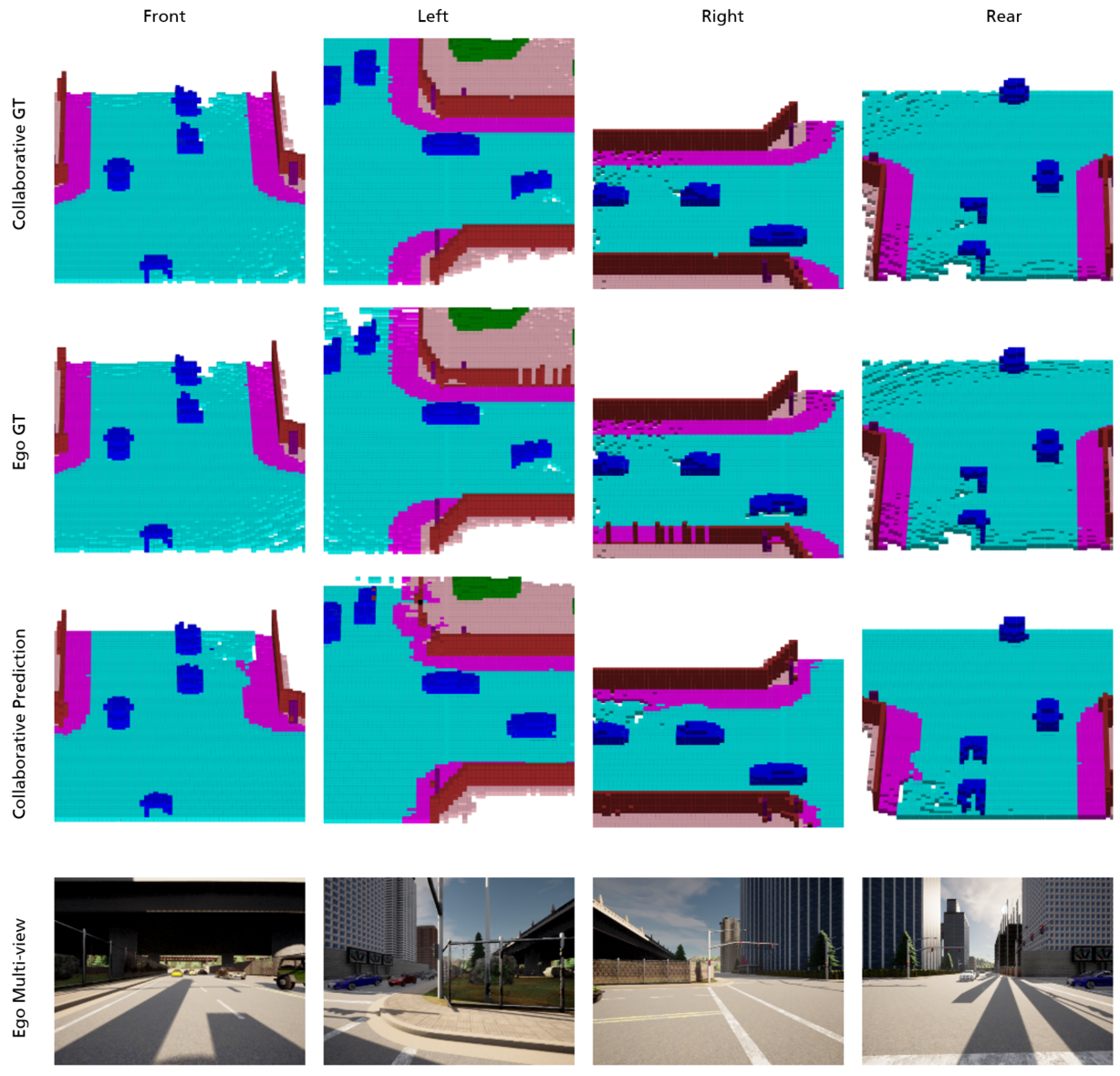}
   \caption{Visual comparison of CoHFF prediction results with collaborative and ego GT in an urban junction scenario.}
   \label{fig:visual_analysis_2}
\end{figure*}

\begin{figure*}[t!]
   \centering
   \includegraphics[trim={0 0 0 0},clip, width=0.98\textwidth]
   {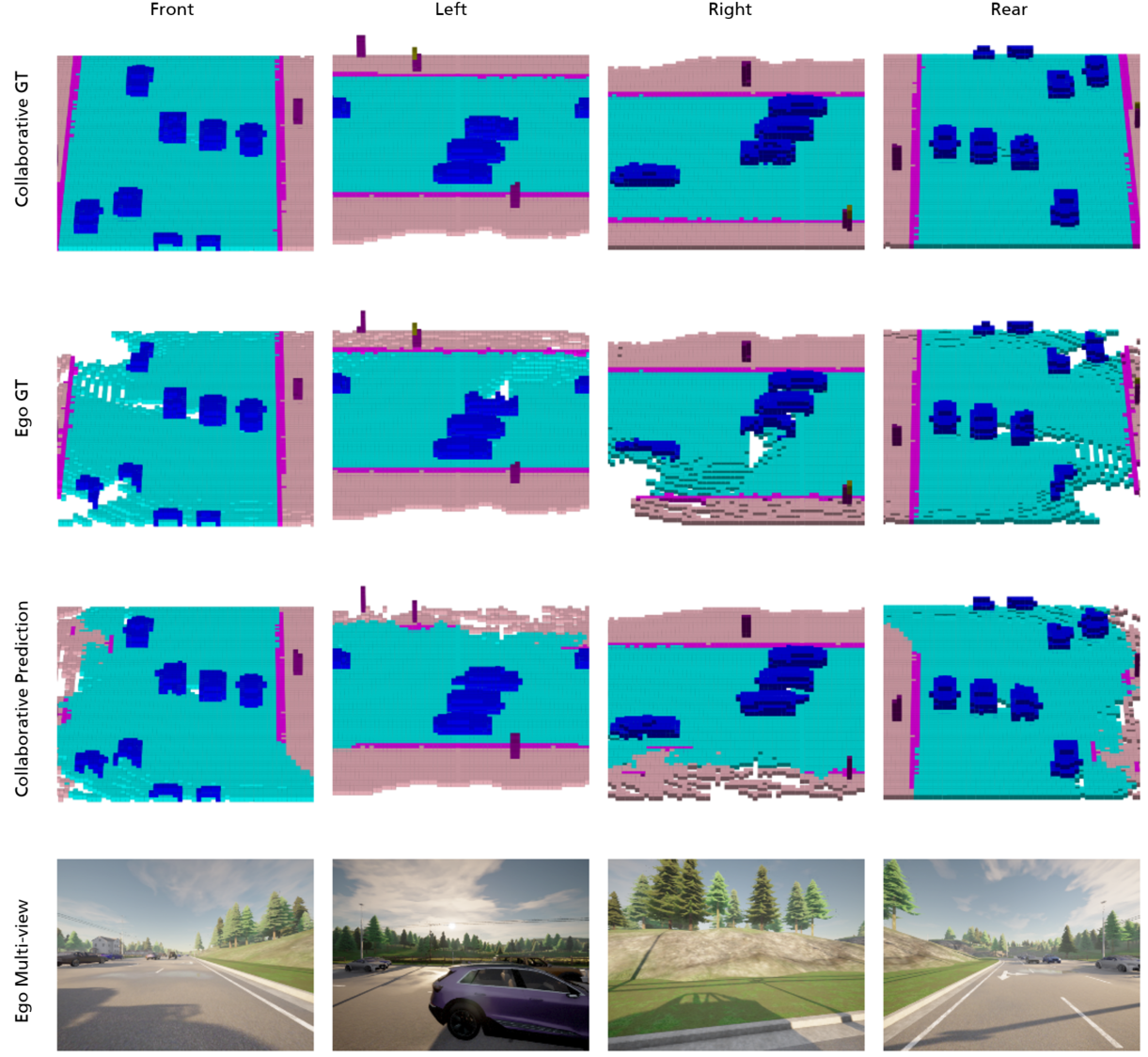}
   \caption{Visual comparison of CoHFF prediction results with collaborative and ego GT on a highway scenario.}
   \label{fig:visual_analysis_3}
\end{figure*}

We provide a further visual comparison of CoHFF prediction results with collaborative and ego ground truth (GT) in an urban lane-change scenario in Fig.~\ref{fig:visual_analysis_1}, an urban junction scenario in Fig.~\ref{fig:visual_analysis_2} and a highway scenario in Fig.~\ref{fig:visual_analysis_3}. Our results demonstrate that the collaborative semantic occupancy prediction using CoHFF can achieve more complete perception than the ground truth in ego GT.

\end{document}